%% file: aaai25.tex
\documentclass[letterpaper]{article} 
\usepackage{aaai25}  
\usepackage{times}  
\usepackage{helvet}  
\usepackage{courier}  
\usepackage[hyphens]{url}  
\usepackage{graphicx} 
\usepackage{natbib}  
\usepackage{caption} 
\usepackage{graphicx}
\usepackage{algorithm}
\usepackage{algorithmic}
\usepackage{amssymb, amsmath}

\usepackage{newfloat}
\usepackage{listings}

\usepackage[utf8]{inputenc} 
\usepackage[T1]{fontenc}    
\usepackage{url}            
\usepackage{booktabs}       
\usepackage{amsfonts}       
\usepackage{nicefrac}       
\usepackage{microtype}      
\usepackage{xcolor}         
\usepackage{tcolorbox}

\title{Textualize Visual Prompt for Image Editing via Diffusion Bridge}

\usepackage{newfloat}
\usepackage{listings}
\DeclareCaptionStyle{ruled}{labelfont=normalfont,labelsep=colon,strut=off} 
\floatstyle{ruled}
\newfloat{listing}{tb}{lst}{}
\floatname{listing}{Listing}
\title{Textualize Visual Prompt for Image Editing via Diffusion Bridge}

\author{
    Pengcheng Xu\textsuperscript{\rm 1,2},
    Qingnan Fan\textsuperscript{\rm 2},
    Fei Kou\textsuperscript{\rm 2},
    Shuai Qin\textsuperscript{\rm 2},
    Hong Gu\textsuperscript{\rm 2},
    Ruoyu Zhao\textsuperscript{\rm 2,3},\\
    Charles Ling\textsuperscript{\rm 1},
    Boyu Wang\textsuperscript{\rm 1 *}
}
\affiliations{
    \textsuperscript{\rm 1} Western University\\
    \textsuperscript{\rm 2} VIVO\\
    \textsuperscript{\rm 3} Xidian University\\
    \{pxu67, charles.ling\}@uwo.ca,
    fqnchina@gmail.com,
    koufei@hotmail.com,
    royzhao@stu.xidian.edu.cn,\\
    \{guhong, shuai.qin\}@vivo.com,
    bwang@csd.uwo.ca\\
}

\usepackage{bibentry}

\begin{document}

\maketitle

\begin{figure*} 
\centering 
\includegraphics[width=0.95\textwidth]{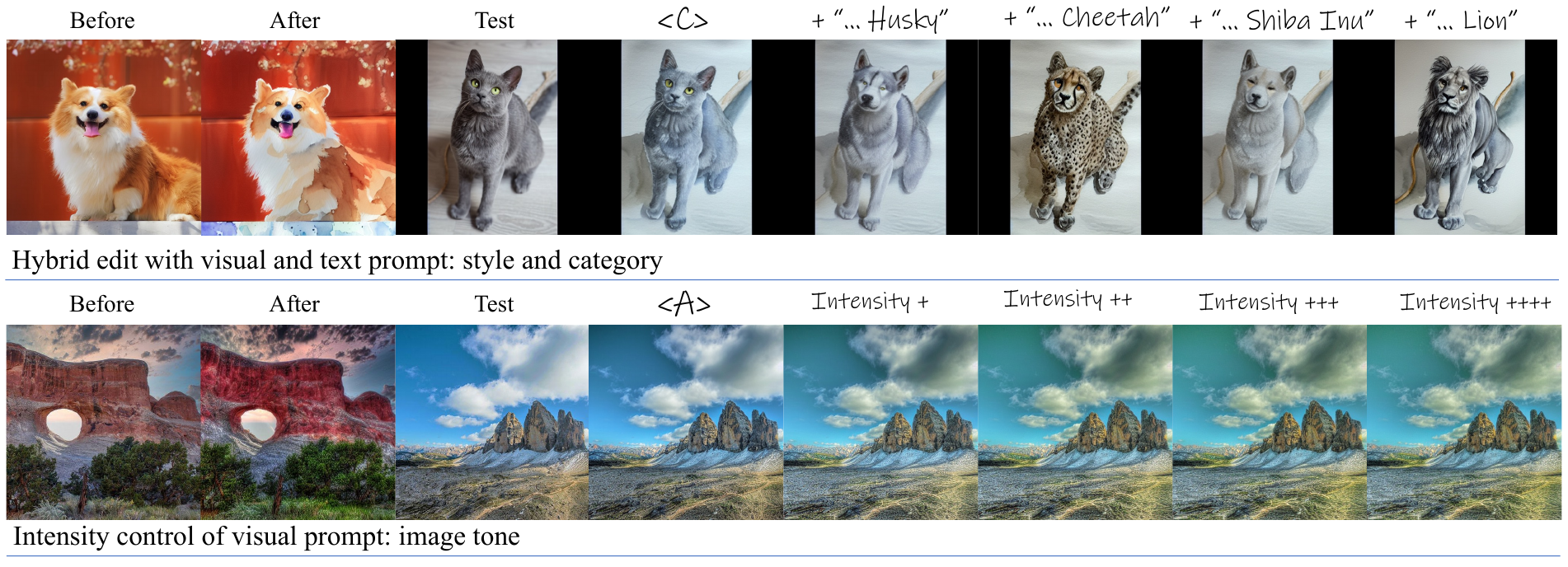} 
\caption{\textbf{Image editing via visual prompt}. The visual prompt defines the visual transformation, which is difficult to describe accurately by language, by a before-and-after image pair. Our method learns such delicate transformation into pseudo text (<A> and <C>), supports hybrid editing with natural text, and can control the intensity of editing with rigorous consistency.} 
\label{pipeline}
\end{figure*}

\begin{abstract}
Visual prompt, a pair of before-and-after edited images, can convey indescribable imagery transformations and prosper in image editing. However, current visual prompt methods rely on a pretrained text-guided image-to-image generative model that requires a triplet of text, before, and after images for retraining over a text-to-image model. Such crafting triplets and retraining processes limit the scalability and generalization of editing. In this paper, we present a framework based on any single text-to-image model without reliance on the explicit image-to-image model thus enhancing the generalizability and scalability.
Specifically, by leveraging the probability-flow ordinary equation, we construct a diffusion bridge to transfer the distribution between before-and-after images under the text guidance. By optimizing the text via the bridge, the framework adaptively textualizes the editing transformation conveyed by visual prompts into text embeddings without other models. Meanwhile, we introduce differential attention control during text optimization, which disentangles the text embedding from the invariance of the before-and-after images and makes it solely capture the delicate transformation and generalize to edit various images. Experiments on real images validate competitive results on the generalization, contextual coherence, and high fidelity for delicate editing with just one image pair as the visual prompt.
\end{abstract}
\begin{links}
\centering
\small
\link{Project page}{pengchengpcx.github.io/TextVDB/}
\end{links}

%

\input{secs/intro}

\input{secs/related}

\input{secs/method1_formulation}

\input{secs/method2_diff_attn}

\input{secs/experiment}

\input{secs/analysis}

\input{secs/conclusion}

\section*{Acknowledgments}
We appreciate constructive feedback from anonymous reviewers and meta-reviewers. This work is supported by Natural Sciences and Engineering Research Council of Canada (NSERC), Discovery Grants program.

\bibliography{aaai25}

\clearpage
\appendix
\input{supp}

\end{document}

%% file: secs/intro.tex
\section{Introduction}

Prompting~\citep{touvron2023llama,floridi2020gpt,liu2023pre}, providing specific instructions or context for the model, is an effective and emergent tool to guide the large-scale text-to-image (T2I) models to generate~\citep{rombach2022high,podell2023sdxl,saharia2022photorealistic,ramesh2022hierarchical} or edit~\citep{hertz2022prompt,kawar2023imagic,parmar2023zero,mokady2023null,gal2022image} remarkable images.
However, in the visual task, not all aspects can be accurately and comprehensively described through language alone. For instance, how to explain the tone transformation between photographs or the personal repaint of a painting? These imagery transformations are abstract, complex, and difficult to convey via a text prompt. In such cases, a pair of before-and-after images, serving as the visual prompt, are more precise and expressive to represent the imagery transformation and guide the image generation and editing. This motivates us to explore the visual prompt for image editing: \textit{given a pair of before-and-after images as a visual prompt, how to distill the transformation from the visual prompt into text embedding for the text-guided image editing}.

Early on, visual prompt for image editing is employed with in-context learning, primarily effective on classical tasks (e.g., segmentation, detection)~\citep{bar2022visual,wang2024context}. Recently, it is explored within diffusion models for arbitrary visual content editing~\citep{nguyen2024visual,motamed2025lego,cheng2023general,yang2024imagebrush,gu2024analogist}, and there are mainly three types of approaches.

The first category relies on the text-guided image-to-image (TI2I) model (e.g., InstructPix2Pix)~\citep{brooks2023instructpix2pix}. Such a TI2I model has an explicit text-guided image-to-image transformation process. By directly optimizing the text condition that guides the model to manipulate and change the before-image to the after-image, the model converts the transformation between the image pair into text~\citep{gal2022image,nguyen2024visual}. However, training such a TI2I model needs to construct a data triplet of text, and before-and-after images. More importantly, such training data is constructed based on the T2I model. For more delicate editing, crafting such triplets requires meticulous effort, limiting the scalability of the training data in the TI2I model~\cite{zhang2024magicbrush}.
Similarly, the second type in-context learning methods~\cite{gu2024analogist,yang2024imagebrush} leverages the diffusion inpainting model which also requires extra complex data (mask and cropped images) for more training to get good generalization ability. This contrasts with the general T2I model, which utilizes a simpler tuple of images and text for training, making it easier to collect and scale up. The third category adopts textual inversion~\citep{gal2022image,motamed2025lego,vsubrtova2023diffusion} to invert the before-and-after images into text embeddings sequentially and learn the transformation or concept by comparison of two embeddings. However, textual inversion is inferior for image reconstruction and cannot capture all image details, making it challenging to learn delicate editing transformations and resulting in a coarse concept. Thus, these designs limit the model to learning precise transformations and generalizing to edit high-fidelity images.

To address the aforementioned issues, in this paper, we aim to answer the following questions: \textbf{why and how to use a general T2I model to learn delicate imagery transformation from visual prompts}. For the why, considering the training data of the TI2I model is mainly derived from the T2I model and manually crafting large-scale triplet (or more complex) data is cumbersome, visual prompts learned directly based on the general T2I model can be more scalable and generalized. For the how, there are two challenges: 1) Construct and textualize the image-to-image (I2I) process but with the single T2I model. The T2I model only has text-to-image mapping but lacks explicit I2I mapping. We have to construct the I2I translation process for the image pair and distill the process into text embedding used for later editing. 2) Capture imagery details for delicate transformation. The constructed I2I process should encode all details of visual prompts otherwise the learned transformation is ambiguous.

To solve these challenges, we propose to textualize the visual prompt based on the diffusion bridge. Firstly, we use a single T2I model to build a diffusion bridge~\citep{zhou2023denoising,su2022dual}, that transforms the distribution of the before-image to the after-image, to represent the I2I process. As shown in Fig.~\ref{pipeline}, the bridge is built on a single pretrained T2I model by leveraging its unconditional (null text) and conditional (text) generating abilities. Based on the probability-flow ordinary equation, the before-image is first transformed into a deterministic latent with the unconditional model and then regenerated to the after-image with the text condition. Such a design breaks the dependency on the explicit I2I process and thus requires no retraining of any TI2I model. Besides, leveraging the more general T2I model, the method adaptively inverts different visual prompts into different text embeddings, supporting generalized, hybrid, and high-fidelity editing.

Secondly, to make the text embedding focus on capturing transformation details, and generalize to edit various images, we propose differential attention control to preserve detailed contents of the before-image during training. This module injects attention weights of the before-image while supporting backpropagation for optimizing text embedding. By learning with attention from the before image, the inverted text embedding is disentangled from the invariance of the before-image, which is irrelevant to the transformation, focuses on capturing fine-grained transformations, and generalizes to edit various images. We summarize our contributions and findings: 
\begin{itemize}
\item[$\bullet$] We design a framework to textualize visual prompts via the T2I diffusion bridge for image editing. The method does not require both image and text conditions as the TI2I model that needs triplets of training data and retraining, which improves the scalability and generalization. 
\item[$\bullet$] We introduce differential attention control for optimizing the text embedding, learning delicate imagery transformations, and generalizing to edit various types of images. 
\item[$\bullet$] Experiments with various evaluation metrics validate our method can learn delicate imagery transformation and generalize better to edit faithful and high-fidelity images, compared to existing methods.
\end{itemize}

%% file: secs/related.tex
\section{Related Work}
\noindent\textbf{Text guided image-to-image models.}
To make the text-to-image (T2I) models control the spatial content of the synthesized images, the text-guided image-to-image model (TI2I) adds an extra image condition based on the T2I model. ControlNet~\citep{zhang2023adding,zhao2024uni} and Adapter~\citep{mou2023t2i,ye2023ip} methods create an additional branch to the UNet to introduce the conditions for generation. Another type of approach combines the embedding of the image with the latent in the diffusion and retrains the model to have both image and text conditions such as InstructPix2Pix~\citep{brooks2023instructpix2pix}, InstructDiffusion~\citep{brooks2023instructpix2pix}, and MagicBrush~\citep{zhang2024magicbrush}. While these TI2I models support human-intuitive text for image editing, collecting image pairs strictly aligned with editing texts is not as easy as simply collecting tuples of the image and its text description, and retraining is inefficient, which limits the capacity of these models. In contrast, our method is directly built on the T2I model for its large text-to-image prior that can be easily scaled up with the increasing scale of pairs of text and image.

\noindent\textbf{Image editing via the visual prompt.} 
Visual prompting was initially proposed in NLP~\citep{brown2020language} based on in-context learning and recently introduced to computer vision~\citep{bar2022visual,wang2023images,wang2023seggpt}. ImageBrush~\citep{yang2024imagebrush} and Analogist~\cite{gu2024analogist} formulate the editing as inpainting in analogy to the image pair but it requires a retrained inpainting model and is limited in image resolutions and encoding intricate details. Image analogy~\citep{vsubrtova2023diffusion,hertzmann2023image} adapts the analogous relation of image pair to new images but the learned editing relation is coarse and ambiguous and cannot accurately edit or preserve structures. Recent methods extract the image transformation from image pairs into text embeddings for text-guided editing. VII~\citep{nguyen2024visual} leverages the pretrained TI2I model, InstructPix2Pix, to distill the image-to-image transformation between the image pair into text embeddings. Lego~\citep{motamed2025lego} and DIA~\citep{hertzmann2023image} uses textual inversion to learn the disentangled concept by comparing the image pair. However, textual inversion is inferior in image reconstruction and cannot encode all imagery details, which causes the learned concept to be coarse and inaccurate. In contrast, our method accurately recovers the image and textualizes the visual prompts without dependence on the pretrained TI2I model, and produces generalized and delicate editing results.

\input{figs/pipeline}

%% file: figs/pipeline.tex
\begin{figure*}[t] 
\centering 
\includegraphics[width=1\textwidth]{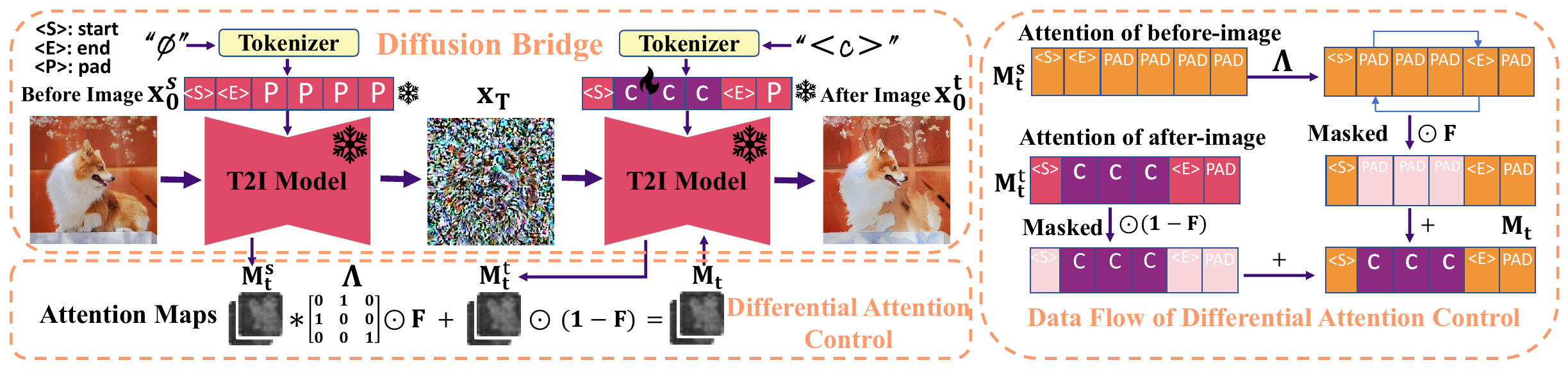} 
\caption{\textbf{Textualization of the diffusion bridge}. \textbf{Left}: The before-image is first transferred to a deterministic latent encoding via the unconditional model and then to the after-image under the text guidance. The text embeddings are optimized with fixed start (latent $\mathbf{x}_T$) and end (after-image $\mathbf{x}_0^a$) states. \textbf{Right}: In training, the attention of the before-image $M_t^b$ is first timed with the column-transformation matrix $\mathbf{\Lambda}$ to switch the column of <E> (end) token, then masked with $\mathbf{F}$. The attention of the after-image $M_t^a$ is masked with $1-\mathbf{F}$ to get the attention of the $y$ tokens. The final $M_t$ is the addition of two masked attentions. This preserves the linguistic format of cross-attention and enables the embedding to learn disentangled and generalized transformation.}
\label{pipeline}
\end{figure*}

%% file: secs/method1_formulation.tex
\section{Methodology}
We aim to learn the delicate imagery transformation from the visual prompt and use it for generalized and high-fidelity image editing. In the following, we first briefly present the score-based generative model~\citep{song2020score} and the diffusion bridge~\citep{song2020denoising,su2022dual} in Sec~\ref{sec:preli}. Then, we discuss converting the visual prompt into text embedding via diffusion bridge in Sec~\ref{sec:inv}. In Sec~\ref{sec:DAC}, we introduce a differential attention control strategy during text optimization, which helps learn disentangled and generalized text guidance. Last, the full algorithm is in Sec~\ref{sec:final}.

\subsection{Preliminaries}
\label{sec:preli}
\noindent\textbf{Scored-based generative model.} \citet{song2020score} proposed the unified diffusion framework with Stochastic Differential Equations (SDE) and further showed that any diffusion process can be represented by a deterministic probability flow (PF) ordinary differential equation (ODE) that transfers the data encodings to \textit{deterministic} and \textit{unqiue} latent encodings. By solving the PF ODE in Eq.~\ref{eq:ode} forward and backward, the data $\mathbf{x}_0$ and latent $\mathbf{x}_T$ are transferred between each other.
\begin{equation}
\small
\label{eq:sde}
\mathrm{d} \mathbf{x}_t=\boldsymbol{\mu}\left(\mathbf{x}_t, t\right) \mathrm{d} t+\sigma(t) \mathrm{d} \mathbf{w}_t
\end{equation}
\begin{equation}
\small
\label{eq:ode}
\mathrm{d} \mathbf{x}_t =\left[\boldsymbol{\mu}(\mathbf{x}_t, t)-\frac{1}{2} \sigma(t)^2 \nabla_{\mathbf{x}} \log p_t(\mathbf{x}_t)\right] \mathrm{d} t
\end{equation}

Here, $\mathbf{w}_t$ is the standard Wiener process, $\boldsymbol{\mu}(t)$ and $\sigma(t)$ are drift and diffusion coefficients respectively. The score function $\nabla_{\mathbf{x}} \log p_t(\mathbf{x}_t)$ is approximated with neural network $\mathbf{s}_\theta(\mathbf{x}_t,t)$. For the T2I model, the text condition $\mathbf{c}$ is added to replace the marginal score function $\nabla_{\mathbf{x}} \log p_t(\mathbf{x}_t)$ with the conditional score function $\nabla_{\mathbf{x}} \log p_t(\mathbf{x}_t | \mathbf{c})$ via classifier-free guidance~\citep{ho2022classifier} in Eq.~\ref{eq:cfg}, where $w$ is the guidance scale, and $\varnothing$ represents the null text.
\begin{equation}
\begin{split}
\small
\label{eq:cfg}
\nabla_{\mathbf{x}} \log p_t(\mathbf{x}_t | \mathbf{c}) & \approx \Tilde{\mathbf{s}}_\theta(\mathbf{x}_t,t,\mathbf{c}) = \mathbf{s}_\theta(\mathbf{x}_t,t, \varnothing) \\ 
&+ w \cdot (\mathbf{s}_\theta(\mathbf{x}_t,t, \mathbf{c}) - \mathbf{s}_\theta(\mathbf{x}_t,t, \varnothing))
\end{split}
\end{equation}

\noindent\textbf{Dual diffusion implicit bridges (DDIB).} Following the above PF ODE in Eq.~\ref{eq:ode}, DDIB~\citep{su2022dual} uses two diffusion models $\mathbf{s}_\theta^{(s)}$ and $\mathbf{s}_\theta^{(t)}$ trained on separated (source and target) domains to perform I2I transformation. Concretely, as shown in Eq.~\ref{eq:ddib}, it first uses an ODE solver in Eq.~\ref{eq:odesolve} to forward the source image $\mathbf{x}^{(s)}$ to the latent $\mathbf{x}^{(l)}$, and then reverse the latent to the target image $\mathbf{x}^{(t)}$. Such a translation process has been theoretically proven to be the most efficient optimal transport between two distributions~\citep{su2022dual}.
\begin{equation}
\small
\label{eq:ddib1}
\mathbf{x}^{(l)}=\operatorname{ODESolve}\left(\mathbf{x}^{(s)} ; \mathbf{s}_\theta^{(s)}, 0,1\right)
\end{equation}
\begin{equation}
\small
\label{eq:ddib}
 \mathbf{x}^{(t)}=\operatorname{ODESolve}\left(\mathbf{x}^{(l)} ; \mathbf{s}_\theta^{(t)}, 1,0\right)
\end{equation}
\begin{equation}
\small
\label{eq:odesolve}
\operatorname{ODESolve}\left(\mathbf{x}_t; \mathbf{s}_\theta, t_0,t_1\right) = \mathbf{x}_{t_0} + \int_{t_0}^{t_1} \mathbf{s}_\theta(\mathbf{x}_t, t) \,\mathrm{d}t
\end{equation}

Our method is driven and designed by such theoretical advantage over transport efficiency. However, instead of using two different models trained on two different domains, we only use a single T2I model and leverage the text condition to transfer to versatile domains.

\subsection{Visual Prompt Learning via Diffusion Bridge}
\label{sec:inv}
\noindent\textbf{T2I diffusion bridge.} To learn the transformation represented by the visual prompt on the more general and scalable T2I model, we need to construct the I2I process (before-to-after image transformation) based on a single T2I model. Inspired by the DDIB, we build a diffusion bridge to represent the I2I transformation. Specifically, we use unconditional (null text  $\varnothing$) and conditional (text $\mathbf{c}$) models to replace the two different diffusions in DDIB, and adopt DDIM as the ODE solver. This can be implemented with a single pretrained T2I model and require no retraining of any model.

Concretely, let $\mathbf{x}_0^b \sim p(\mathbf{x}^b)$ and $\mathbf{x}_0^a \sim p(\mathbf{x}^a)$ represent the before-and-after images, respectively. The intermediate latent is $\mathbf{x}_T \sim \mathcal{N}(0, I)$. Our diffusion bridge is defined in Eq.~\ref{eq:ddim_inv} and \ref{eq:ddim} in which the before-image $\mathbf{x}_0^b$ is first transformed into $\mathbf{x}_T$ with null text $\varnothing$ in $T$ steps and then regenerated to $\mathbf{x}_0^a$ under the guidance of text embedding $\mathbf{c}$. Such a process model the distribution transition of $p(\mathbf{x}^b) \to p(\mathbf{x}^a)$. We aim to learn $\mathbf{c}$ that can guide any samples drawn from $p(\mathbf{x}^b)$ to $p(\mathbf{x}^a)$ in a few-shot manner.
\begin{equation}
\small
\label{eq:ddim_inv}
\mathbf{x}_0^b \longrightarrow \mathbf{x}_T: \mathbf{x}_T =\operatorname{DDIM}\left(\mathbf{x}^b_0 ; \Tilde{\mathbf{s}}_\theta(\mathbf{x}_t,t,\varnothing), 0,T\right)
\end{equation}
\begin{equation}
\small
\label{eq:ddim}
\mathbf{x}_T\longrightarrow \mathbf{x}_0^a: \mathbf{x}^a_0 =\operatorname{DDIM}\left (\mathbf{x}_T ; \Tilde{\mathbf{s}}_\theta(\mathbf{x}_t,t, \mathbf{c}), T,0\right)
\end{equation}

Note that the DDIM in Eq.~\ref{eq:ddim_inv} and ~\ref{eq:ddim} is deterministic which means that once the prior model $\Tilde{\mathbf{s}}_\theta$ is determined, for any $\mathbf{x}^b_0 \to \mathbf{x}_T$ in Eq.~\ref{eq:ddim_inv}, $\mathbf{x}_T$ is \textit{deterministic} and \textit{unique}. The same applies to $\mathbf{x}_T \to \mathbf{x}^a_0$. This forms a unique one-to-one mapping for every pair of before-and-after images, which enables high-fidelity and stable image editing.

However, such a property makes the widely used textual inversion~\citep{gal2022image} invalid. This is because when optimizing the text embedding $\mathbf{c}$, textual inversion adds random noise to the image at every step, and the final latent $\mathbf{x}_T$ will not be deterministic anymore, which violates our constraint for $\mathbf{x}_T$. Essentially, the learning process of textual inversion tries to map a single image $\mathbf{x}^a_0$ to the whole latent Gaussian distribution rather than the fixed noise vector $\mathbf{x}_T$. Apart from this, this stochastic procedure is inferior in image reconstruction and results in the learned text embeddings being unable to capture and recover all imagery details~\citep {mokady2023null}. Mostly, textual inversion is effective in learning objects but not the whole image and details, and always introduces significant changes. As a result, it is not suitable for image editing~\citep{nguyen2024visual}. This motivates us to design the optimization procedure of the text embedding $\mathbf{c}$ with fixed start $\mathbf{x}_T$ and end $\mathbf{x}_0^a$ states. 

\noindent\textbf{Optimization conditioned on both start and end states of diffusion.} The text embedding $\mathbf{c}$ is optimized to satisfy the diffusion process $\mathbf{x}_T \to \mathbf{x}^a_0$ given the start and end states $\{\mathbf{x}_T, \mathbf{x}_0^a\}$. Such textualization based on the implicit bridge is conditioned on both the start and end states. Thus, we optimize the text embedding $\mathbf{c}$ to maximize the conditional probability $p_{\mathbf{c}, \theta} (\mathbf{x}^a_0 | \mathbf{x}_T) $ which is parameterized by the T2I model. This is different from the general diffusion process in textual inversion~\citep{gal2022image} whose end state $\mathbf{x}_T$ is an \textit{unconstrained} Gaussian noise. We discuss the optimization design of our textualization as follows. With the deterministic DDIM, the process in Eq.~\ref{eq:ddim} for each timestep $t$ is written as:
\begin{equation}
\small
\label{eq:opt}
\mathbf{x}_{t-1} = \sqrt{\alpha_{t-1}} \mathbf{f}_\theta( \mathbf{x}_{t}, t, \mathbf{c}) + \sqrt{1-\alpha_{t-1}} \Tilde{\mathbf{s}}_\theta(\mathbf{x}_t,t, \mathbf{c})
\end{equation}
\begin{equation}
\small
\label{eq:opt2}
\mathbf{f}_\theta( \mathbf{x}_{t}, t, \mathbf{c} ) = \frac{\mathbf{x}_{t} - \sqrt{1-\alpha_{t}} \Tilde{\mathbf{s}}_\theta(\mathbf{x}_t,t, \mathbf{c})} {\sqrt{\alpha_t}}
\end{equation}
Considering that the start and end states $\{\mathbf{x}_T, \mathbf{x}_0^a\}$ are given and fixed and the DDIM is deterministic, one intuitive way to optimize $\mathbf{c}$ is to first do forward pass $\mathbf{x}_T \to \mathbf{x}^a_0$ to get the final output image $\mathbf{\hat{x}}_0$ with Eq.~\ref{eq:opt} and calculate the loss with $\| \mathbf{\hat{x}}_0 - \mathbf{x}^a_0 \|_2$. However, this recurrent backpropagation needs to cache the intermediate result of $\mathbf{s}_\theta$ at each timestep $t$, which is not feasible for multiple steps in diffusion.

To cope with this, we propose to optimize the \textit{predicted} $\mathbf{x}_0$, $\mathbf{f}_\theta( \mathbf{x}_{t}, t )$, with the ground-truth after-image $\mathbf{x}_0^a$. The predicted $\mathbf{x}_0$, $\mathbf{f}_\theta( \mathbf{x}_{t}, t )$ is an estimator of the final output at the current timestep, which indicates how far away the current result is from the desired output~\citep{song2020denoising}. Besides, we find that the losses in the initial steps are much larger than those in the final steps. Optimizing the loss at each timestep equally may cause $\mathbf{\hat{x}}_0$ to deviate from $\mathbf{x}_0^a$ at the final step. To make the last step $\mathbf{\hat{x}}_0 \approx \mathbf{x}_0^a$, we scale down the loss with time-aware scaling function $\beta(t)$ and the optimization at each timestep is:
\begin{equation}
\small
\label{eq:obj}
\mathcal{L}(\mathbf{c}, t) = \beta(t) \|\mathbf{f}_\theta( \mathbf{x}_{t}, t, \mathbf{c}) -  \mathbf{x}_0^a\|_2 
\end{equation}

%% file: secs/method2_diff_attn.tex
\subsection{Learning with Differential Attention Control}
\label{sec:DAC}
Attention injection can preserve the invariance of before-and-after editing, which enables saving the before-image content in image editing. It is generally used in \textit{inference} but not in \textit{training} the T2I model. Besides, current attention-based image editing methods~\citep{hertz2022prompt,cao2023masactrl,tumanyan2023plug} are not straightforward differential and do not support backpropagation.
%


\noindent\textbf{Learning motivation.} We introduce attention injection during \textit{training} the text embedding $\mathbf{c}$. Intuitively, the injected attention introduces the information from the before-image and thus disentangles the learned text embedding $\mathbf{c}$ from the before-image information. This makes the text embedding $\mathbf{c}$ solely learn the transformation and generalize to edit various images. Otherwise, the information irrelevant to the transformation from the before-image leaks to $\mathbf{c}$ such that it only fits on the before-image but cannot generalize on other images. In summary, our motivation for training with attention injection has two aspects: 1) In training, leveraging the injected attention capturing the invariance between the before-and-after images, the text embedding learns a visual transformation disentangled from a specific before-image and can be more generalized. 2) In inference, the injected attention preserves the invariance of the before-image and achieves high fidelity.

\noindent\textbf{Module design.} To make the attention injection differential and support backpropagation, we implement the attention injection with only multiplication and addition. The detailed process is depicted in Fig.~\ref{pipeline}. For simplicity, let $M_t^b, M_t^a \in \mathbb{R}^{j \times k}$ denote the attention weights from before-and-after images at timestep $t$, respectively, where $j$ is the feature dimension and $k$ is the length of total tokens. The text prompt $\mathbf{c}$ of the after-image includes $y$ tokens for learning the transformation. So, we keep the attentions of $y$ tokens in $M_t^a$ and replace the rest $k-y$ with attentions from $M_t^b$. 
Besides, we make the $(y+1)^{th}$ attention in $M_t^a$ to be the <end> token attention of $M_t^b$ so that the new attention $M_t$ also follows the linguistic format determined by the text encoder. We use a column transformation matrix $\mathbf{\Lambda}$ and mask $\mathbf{F} =[1,...,0,1,1] \in \{0,1\}^k$ to achieve this.
Specifically, the column-transformation matrix is a modified identity matrix that switches the columns of $1$ and $y+1$. It shifts the cross-attention of the <end> token in $M_t^b$ to the $(y+1)^{th}$ column, and the mask $\mathbf{F}$ injects all cross-attention weights of $M_t^b$ that
are not in the position of tokens of $\mathbf{c}$, into $M_t$.
For self-attention, we do not include $\mathbf{\Lambda}$ since self-attention is not related to text. Consequently, the training objective in Eq.~\ref{eq:obj} is added with the new attention and becomes Eq.~\ref{eq:obj2}. $\odot$ is column-multiplication that multiplies each column in the attention matrix with a scalar value of 0 or 1.
\begin{equation}
\small
\label{eq:attn}
M_t = M_t^b \mathbf{\Lambda} \odot \mathbf{F} +  M_t^a \odot (1-\mathbf{F})
\end{equation}
\begin{equation}
\small
\label{eq:obj2}
\mathcal{L}(\mathbf{c}, t) = \beta(t) \|\mathbf{f}_\theta( \mathbf{x}_{t}, t, \mathbf{c}, M_t) -  \mathbf{x}_0^a\|_2 
\end{equation}

\begin{algorithm}[t]
\small
    \caption{Textualize Visual Prompt}
    \label{algo:optimization}
    \begin{algorithmic}[1]
    \STATE \textbf{Input}: An image pair $\{\mathbf{x}^b_0, \mathbf{x}^a_0\}$
    \STATE \quad T2I diffusion model $\mathbf{s}_\theta$
    \STATE \quad Number of training epochs $N$; Number of diffusion steps $T$
    \STATE \quad Learning rate $\gamma$; Attention injection timestamp $\tau$
    \STATE \quad Initialize $\mathbf{c}$ 
    \FOR{$i=1,\cdots,N$}
        \STATE $\mathbf{x}_T =\operatorname{DDIM}\left(\mathbf{x}^b_0 ; \Tilde{\mathbf{s}}_\theta(\mathbf{x}_t,t,\varnothing), 0,T\right)$
    \FOR{$t=T,\cdots,1$}
        \IF{$t < \tau$}
            \STATE Build column-transformation matrix $\mathbf{\Lambda}$ and mask $\mathbf{F}$
            \STATE Get $M_t$ with attention injection with Eq.~\ref{eq:attn}
        \ENDIF
        \STATE Calculate predicted $\mathbf{x}_0$, $\mathbf{f}_\theta( \mathbf{x}_{t}, t, M_t)$ with Eq.~\ref{eq:opt2}
        \STATE Calculate next state $\mathbf{x}_{t-1}$ with Eq.~\ref{eq:opt}
        \STATE Calculate $\mathcal{L}(\mathbf{c}, t) = \beta(t) \|\mathbf{f}_\theta( \mathbf{x}_{t}, t, \mathbf{c}, M_t) -  \mathbf{x}_0^a\|_2 $ with Eq.~\ref{eq:obj2}
        \STATE Update $\mathbf{c} = \mathbf{c} - \gamma \nabla\mathcal{L}(\mathbf{c}, t)$
    \ENDFOR
    \ENDFOR
\STATE \textbf{Output}: $\mathbf{c}$
\end{algorithmic}
\end{algorithm}

\subsection{Visual Prompt for Image Editing} 
\label{sec:final}
Finally, by constructing a diffusion bridge with the T2I model, we distill the I2I transformation defined by an image pair $\mathbf{x}_0^b$ and $\mathbf{x}_0^a$ into a text embedding $\mathbf{c}$. Our framework is in Algorithm~\ref{algo:optimization}. In inference, the new image goes through the diffusion bridge as the before-image under the guidance of the learned text embedding. The output is the edited image with the desired transformation. Our method also adapts to multiple image pairs with the same transformation. For each epoch, we randomly select an image pair for training.

\begin{figure*}[!htb]
\centering 
\includegraphics[width=0.82\textwidth]{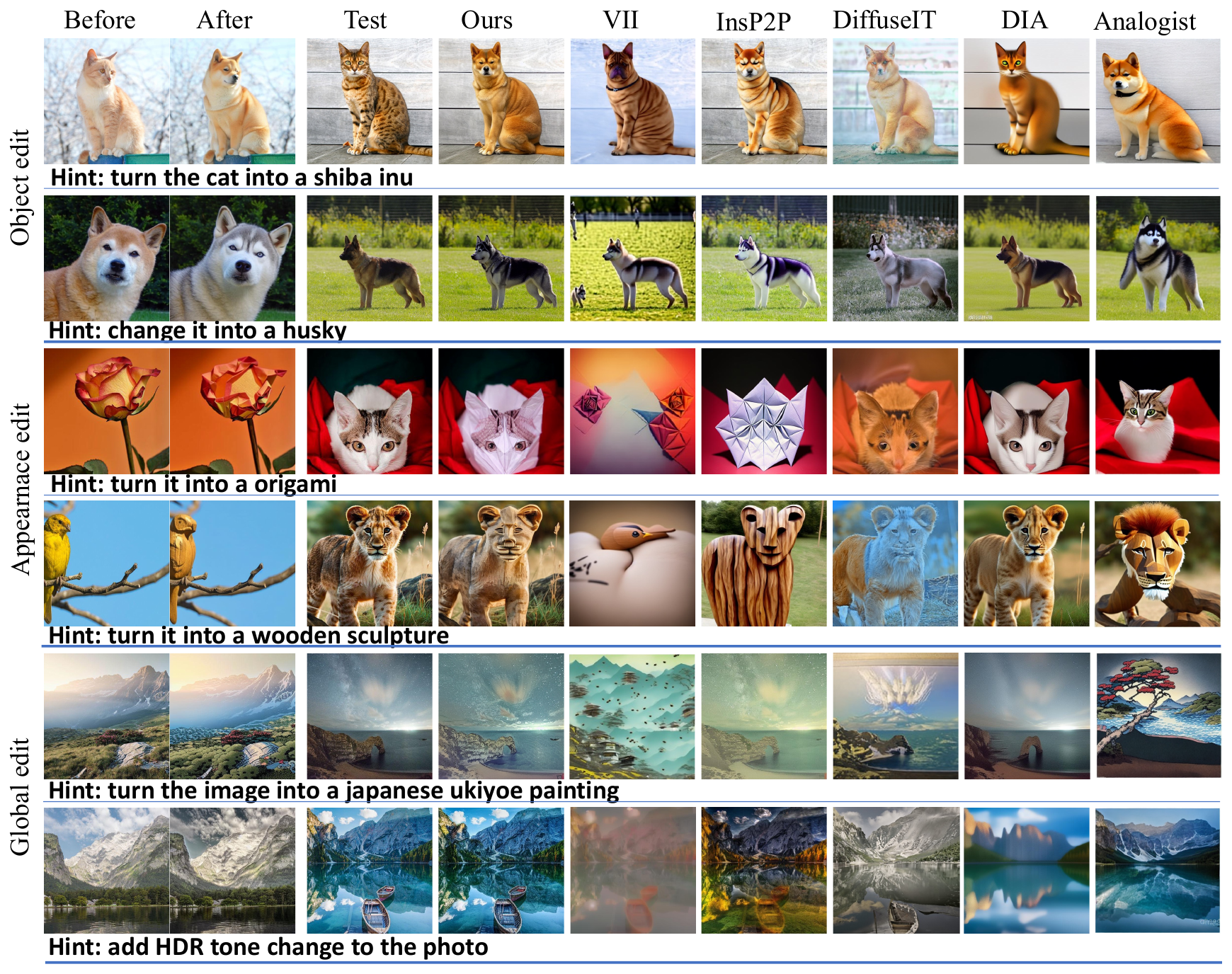} 
\caption{\textbf{Qualitative comparisons on real images}. Visual prompts with different editing types and different levels of geometric changes. Our method generalizes to different editing types and scenes while preserving different levels of geometric structures.}
\label{com}
\end{figure*}

%% file: secs/experiment.tex
\section{Evaluation on Real Images}

\subsection{Experiment setup}
\label{sec:setup}
\noindent\textbf{Baselines and implementations.} The closest work to ours is visual instruction inversion (VII)~\citep{nguyen2024visual} which also learns image transformation by an image pair. VII is based on the pretrained TI2I model, InstructPix2Pix (InsP2P)~\citep{brooks2023instructpix2pix}. We also add InsP2P with ground-truth text instructions, which are actually unavailable. Additionally, we add diffusion image analogy DIA~\citep{vsubrtova2023diffusion} that transfers the analogy relation between the image pair to new images, and visual in-context learning Analogist~\citep{gu2024analogist} that inpainting the edited image with in-context learning, and style transfer methods DiffuseIT~\citep{kwon2022diffusion} that transfer the style of after-image to the new image. We implement our method based on SD v1.5, consistent with VII and InsP2P. All baselines are evaluated with their official codebases.

\noindent\textbf{Datasets.} We evaluate our framework on \textit{real images} based on the two latest benchmarks Dreambooth~\citep{ruiz2023dreambooth} and PIE~\citep{ju2023direct}. Following the procedure of creating image pairs in InsP2P and VII, we also use P2P~\citep{hertz2022prompt} to generate different image pairs with different text instructions as visual prompts. We use this dataset to evaluate the generalization and fidelity of our method since the data is not included in the training set of SD v1.5 or InsP2P. The details of implementation and dataset are in the Appendix~\ref{impl}.

\subsection{Comparion with previous methods}
\label{sec:compa}



\noindent\textbf{Real image editing and fidelity.}
We evaluate our method in real image editing. As shown in Fig.~\ref{com}, our method produces higher fidelity and preserves better invariant details of the original image under different editing types. This indicates that our method can learn more accurate and generalized editing effects from the image pair. In contrast, style transfer (DiffuseIT) and image analogy (DIA) methods cannot disentangle the delicate transformation from irrelevant image contents, leading to ambiguous and coarse results. VII and Analogist cannot preserve the structural and invariant details of the original image. This is unacceptable for image tone transformation such as HDR, that has strict fidelity requirements. With the ground-truth editing instructions, the results of InsP2P are low-fidelity and do not exactly follow the desired pattern in visual prompts. This shows the necessity of visual prompts for indescribable visual transformation.

\begin{figure}[htb] 
\centering 
\includegraphics[width=0.47\textwidth]{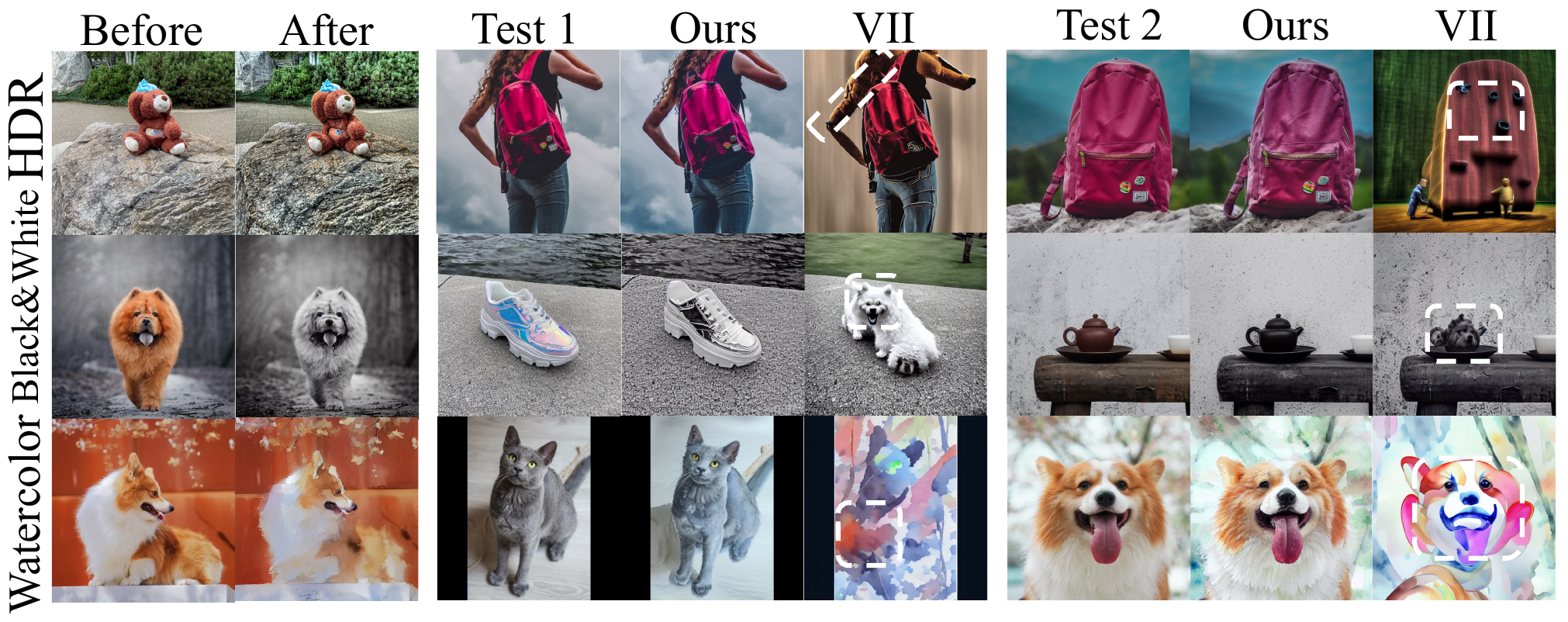}
\caption{\textbf{Generalization in heterogeneous scenes and categories.} Results of tone and style editing show our method does not introduce \textit{leaked content (bear texture, dog face, color)} from visual prompts when the category and scene of test images differ greatly from the visual prompts.} 
\label{tone}
\end{figure}

\noindent\textbf{Leakage of irrelevant content and disentanglement.}
Our method aims to learn a disentangled editing effect by training with attention injection. The learned text embedding should not encode irrelevant information from visual prompts. The results show that our method only learns the editing effect but does not introduce the color, objects, or textures from the visual prompts. Concretely, Fig.~\ref{tone} and~\ref{com} show that VII introduces the bear and its texture (first row), dogs (second row), and color (third row) to the edited images. DiffuseIT learns the transformation only from the single after-image and cannot learn the contrasted and accurate editing from the image pair. DIA cannot accurately learn the disentangled effect from the image pair, yielding inaccurate results.

\noindent\textbf{Generalization.}
Fig.~\ref{tone} also validates the generalization in heterogeneous scenes and categories. Our method can edit images of various scenes and categories with high fidelity whereas VII (based on the TI2I model) degrades significantly when test images' scenes or categories differ greatly from visual prompts. This also validates our motivation to use T2I model with larger prior and differential attention control.

\noindent\textbf{Necessity of visual prompts and ambiguity of text prompt.} We demonstrate the value of visual prompts from two sides. First, some editing effects cannot be accurately \textit{identified or explained} by observation. Second, even if the effect can be identified as text such as `watercolor', the text prompt can be ambiguous, and the `watercolor' defined by the text prompt can be different from the `watercolor' defined in the visual prompt. We demonstrate this in Fig.~\ref{supp_qua2} in Appendix. There are various sub-styles of painting. The `psychedelic painting' is difficult to recognize and the `watercolor' effect in the visual prompt differs from that in InsP2P using text.

\begin{table*}[t]
\centering
\small
\tabcolsep=0.07cm
\scalebox{1}{
\begin{tabular}{@{}l|ccc|cc|cc|cc|ccc@{}}
\toprule
\textbf{Method} & \textbf{PSNR}$\uparrow$ & \textbf{SSIM}$\uparrow$ & \textbf{LPIPS}$\downarrow$ & \textbf{V-CLIP}$\uparrow$  & \textbf{I-CLIP}$\uparrow$ &\textbf{V-DINO}$\uparrow$  &\textbf{I-DINO}$\uparrow$ & \textbf{V-VIE}$\uparrow$  & \textbf{I-VIE}$\uparrow$   &\textbf{Edit Analogy}$\uparrow$ &\textbf{Fidelity}$\uparrow$  &\textbf{Overall}$\uparrow$ \\ \midrule
VII    &12.76 &0.4460 &0.5238 &0.1819 &0.7012 &0.1564 &0.7144 &1.92 &1.69 &3.21 &1.33 &2.27  \\ \midrule
InsP2P  &15.75 &0.5977 &0.3168 &0.2674  &0.8114 &0.2247 &0.8763 &3.01  &2.93    &4.21 &2.93 &3.57   \\ \midrule
DiffuseIT  &12.06 &0.3610 &0.5261 &0.1103 &0.7180 &0.1167 &0.7538 &2.11  &3.32  &2.98 &3.28 &3.13   \\ \midrule
DIA     &17.78  &0.5344  &0.4279  &0.1201  &0.7453 &0.1010  &0.7429 &1.18 &4.01 &2.14 &3.87 &3.01 \\ \midrule
Analogist    &15.26  &0.5102  &0.4061  &0.2123  &0.7037 &0.1914  &0.7419 &2.51  &3.24 &3.98 &3.02 &3.50   \\ \midrule
\textbf{Ours}  &\textbf{24.57} &\textbf{0.8091} &\textbf{0.1197} &\textbf{0.2750}  &\textbf{0.8178} &\textbf{0.3234} &\textbf{0.9073} &\textbf{4.48}  &\textbf{4.29}  &\textbf{4.81} &\textbf{4.62} &\textbf{4.72}  \\ \bottomrule
\end{tabular}
}
\caption{\textbf{Quantitative comparison and user study.} We evaluate editing direction and fidelity in classical metrics, CLIP, DINO, GPT-4o, and human preference. We achieve better editing consistency aligned with the visual prompts and fidelity on all metrics.}
\label{quan}
\end{table*}

\noindent\textbf{Quantitative comparison and user study.}
We extend evaluating metrics in ~\citet{nguyen2024visual,ju2023direct} with DINOv2 and VIE~\cite{ku2023viescore} based on GPT-4o. The VisualCLIP (V-CLIP) measures the cosine similarity of the editing direction between the before/after example pair and test pair based on their CLIP embeddings, indicating the agreement of the editing direction with the visual prompt. The ImageCLIP (I-CLIP) measures the cosine similarity between the edited and original images, indicating the fidelity to the original test image. The same applies to DINOv2. Similarly, A user study evaluates the preservation of test image invariance (Fidelity) and the analogy of the edit direction (Edit Analogy). VIE scores, evaluated by GPT-4o, assess these properties similarly to human criteria~\cite{ku2023viescore}. Our method achieves competitive results across all metrics, consistent with qualitative results. See Appendix~\ref{impl}.

%% file: secs/analysis.tex
\subsection{Ablation and Analysis}
\label{sec:abla}

\begin{figure}
\centering
\includegraphics[width=0.45\textwidth]{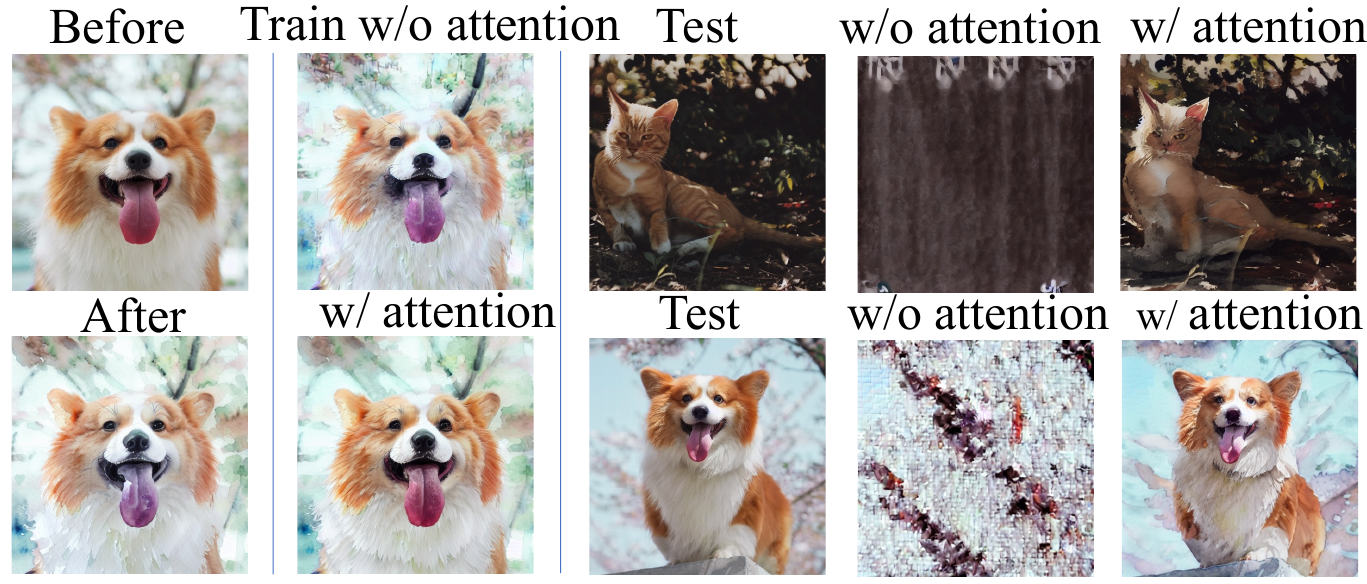} 
\caption{\textbf{Train and test results with/without attention}. 1st column: the visual prompt; 2nd column: training results without/with injected attention. 3rd column: test images; 4\&5 columns: test results without and with attention control.} 
\label{dac_abla}
\end{figure}

\noindent\textbf{Differential attention control.} 
The differential attention control injects the attention of the before-image into that of the after-image. We validate its two benefits with experiments in Fig.~\ref{dac_abla}. First, during training, the injected attention preserves the invariance making the text embedding fit the after-image better. As shown in Fig.~\ref{dac_abla} 2nd column, the model produces a detailed dog portrait with injected attention, unlike the coarse version without it. This helps the text embedding focus on the transformation, improving generalization. Second, during testing, the module can preserve the invariance from the before-image, achieving high fidelity. In 4\&5 columns, the attention module allows the text embedding to edit different images accurately. Without it, outputs are distorted, indicating overfitting to the training image pair.

\begin{figure}
\centering
\includegraphics[width=0.46\textwidth]{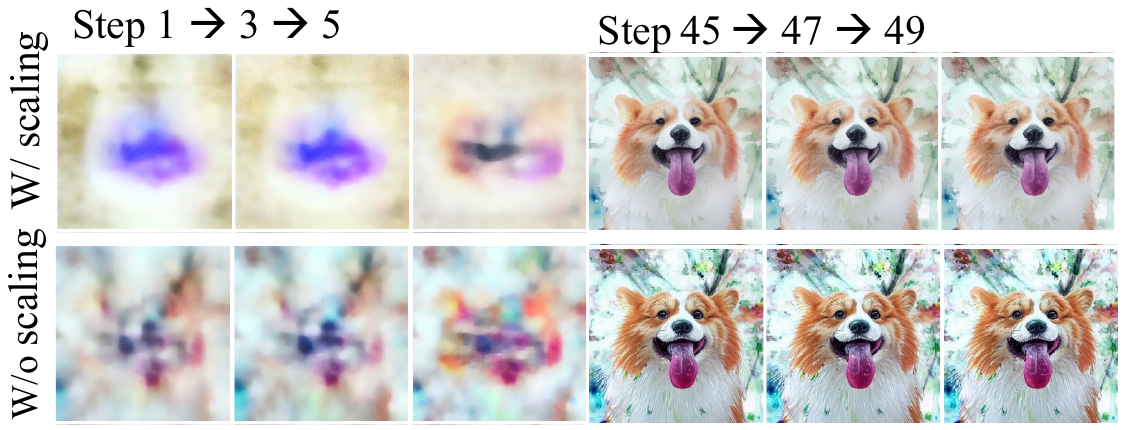} 
\caption{\textbf{The predicted $\mathbf{x}_0$ at different timesteps}. The results of early steps without scaling are closer to $\mathbf{x}_0^b$ but the final output deviates. With scaling, the results of early steps are farther but the final output approaches $\mathbf{x}_0^b$.} 
\label{scale}
\end{figure}

\noindent\textbf{Time-aware scaling loss.} We analyzed the predicted $\mathbf{x}_0$, $\mathbf{f}_\theta( \mathbf{x}_{t}, t )$ along all timesteps. The observation in Fig.~\ref{scale} shows that equally optimizing the loss at each timestep may reduce the loss of predicted $\mathbf{x}_0$ in the early steps too much and make the final output deviate from the ground truth $\mathbf{x}_0^a$. Meanwhile, the output that is closest to $\mathbf{x}_0^a$ appears earlier and the image quality degrades undesirably. After scaling, the final output becomes closest to $\mathbf{x}_0^a$, and the image quality also improves.

\noindent\textbf{Limitation.} 
One limit is the prior of the T2I model. Despite using the more general T2I model with a larger prior, editing real images remains challenging since many practical transformations are out of the model's prior. Learning visual prompts is essentially a process of extracting and composing the prior of the text-to-image. Such prior restricts our method to learning visual prompts and editing for real images. Another is choosing the timestamp of the attention injection for different editing types since the invariance of them is different. To fit the delicate transformation, we need to choose the appropriate timestep for each editing type in training.

%% file: secs/conclusion.tex
\section{Conclusion}
This paper introduces a method using a single, general text-to-image model as a diffusion bridge for learning complex image transformations from visual prompts. This approach removes the dependence on a text-guided image-to-image model and extensive triplet data of text and before-and-after images, which are harder to create than simple text-image pairs and are not scalable on a large scale. We also demonstrate that training with differential attention allows the visual transformation to be embedded into the text space, disentangling the text embedding from the irrelevant before-image information and enhancing the generalizability of text embeddings for various image edits. Results show that leveraging the scalable T2I model achieves high-fidelity image editing with visual prompts, suggesting a new paradigm for leveraging scalable generative models for detailed visual prompts.

%% file: supp.tex
\onecolumn
\section{Supplementary}
\input{secs/supp_qualitative}

\input{secs/supp_quantitative}

%% file: secs/supp_qualitative.tex
The supplementary is arranged as follows: B. Additional related work. C. Additional analysis of the textualization process and the learned text embeddings. D. The analysis of predicted $\mathbf{x}_0$ during optimization. E. The evaluation details for quantitative analysis and user study. F. Additional qualitative editing results and applications.

\section{Additional related work}
\noindent\textbf{Text-to-image models.}
Leveraging the large-scale image-text pairs, recent diffusion-based text-to-image (T2I) models demonstrate new state-of-the-art in image quality and diversity~\citep{nichol2021glide,rombach2022high,podell2023sdxl,saharia2205photorealistic} compared with previous GAN~\citep{reed2016generative,zhang2017stackgan,xu2018attngan,li2019controllable} and autoregressive models~\citep{ramesh2021zero,ding2021cogview,wu2022nuwa,yu2022scaling,esser2021taming} which suffer from small-scale data and large computation cost, respectively. Several works~\citep{li2024blip,lian2023llm} are further developed to enhance the richness and alignment of image and text semantics by integrating with large-scale pretrained models. Our method is directly built on the T2I model for its large text-to-image prior that can be easily scaled up with the increasing scale of text and image pairs.

\noindent\textbf{Diffusion-based image editing.} 
Image editing manipulates visual content without changing others. There are multiple ways to condition the original image during editing. The TI2I model above explicitly adds the origin image features to the model as the condition. Attention-based methods, such as P2P~\citep{hertz2022prompt}, MasaCtrl~\citep{cao2023masactrl}, P2P-Zero~\citep{parmar2023zero}, and PnP~\citep{tumanyan2023plug}, copy the attention weights from the original image during editing. \citet{liu2024towards} further reveal the role of attention within UNet in image generation and editing. \citet{lin2024text,huang2024smartedit,sheynin2024emu} explore more precise and versatile image editing with multi-task learning and Multimodal Large Language Models (MLLMs). SDEdit~\citep{meng2021sdedit} uses the intermediate noisy image to regenerate the edited image. SEGA~\citep{brack2024sega} explores the properties of isolated semantics of the latent space in SD. Fine-tune methods~\citep{kawar2023imagic,kim2022diffusionclip,valevski2022unitune} that only fit a specific image and text prompt, cannot generalize to new images and are inefficient. Our method learns the visual prompts only once and then can accurately edit images without tuning or extra modules. 

\section{Semantic analysis of the learned text embedding}
\begin{figure*}[htb] 
\centering 
\includegraphics[width=0.87\textwidth]{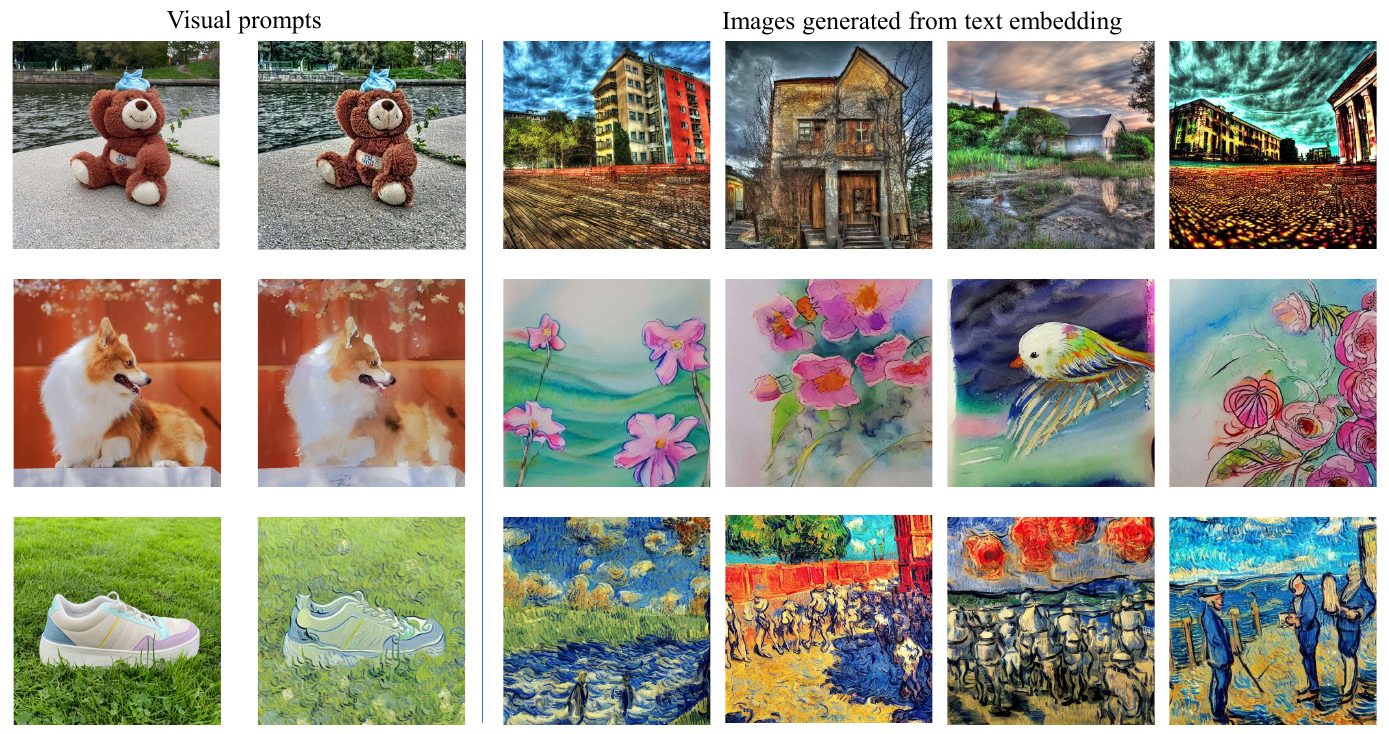} 
\caption{\textbf{Generation results of text embedding.} We use the learned text embeddings of different visual prompts to generate images under different random seeds to visualize the learned transformation concept.} 
\label{abla_gen}
\end{figure*}

We analyzed the learned semantic meaning of the text embedding from different visual prompts. To visualize the semantic meanings of text embeddings, we use solely learned text embeddings to generate images under different random seeds to demonstrate the semantic meanings. As shown in Fig.~\ref{abla_gen}, the style of images generated by the text embeddings is similar to the semantic meaning of the visual transformation. For example, the transformation in the first row is \textit{image tone} transformation and generated images show a similar tone style. The second and third rows are the same. The generated images show the \textit{watercolor} and \textit{van-gogh} painting style. This shows that our learned text embedding successfully captures the semantic meaning of the visual transformation.

\clearpage
\section{Analysis of predicted $\mathbf{x}_0$ in training}
\begin{figure*}[htb] 
\centering 
\includegraphics[width=0.72\textwidth]{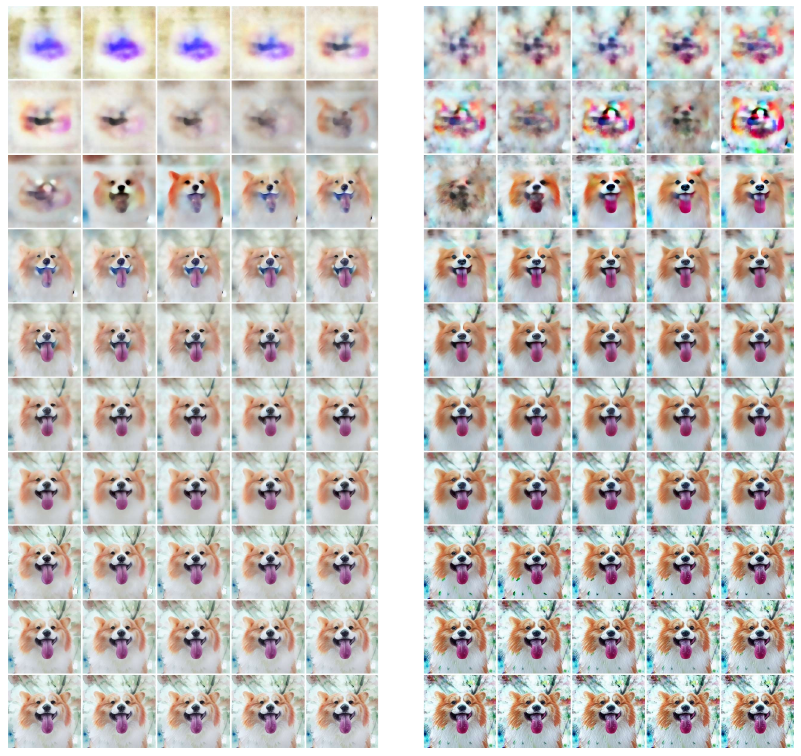} 
\caption{\textbf{Visual comparison the predicted $\mathbf{x}_0$ in all time steps.} We show the predicted $\mathbf{x}_0$ in all timesteps with (left) and without (right) time-aware scaling. From left to right and top to bottom, the time step gradually increases from 0 to 50.}
\label{supp_predx}
\end{figure*}

\begin{figure}[!htb] 
\centering 
\includegraphics[width=0.35\textwidth]{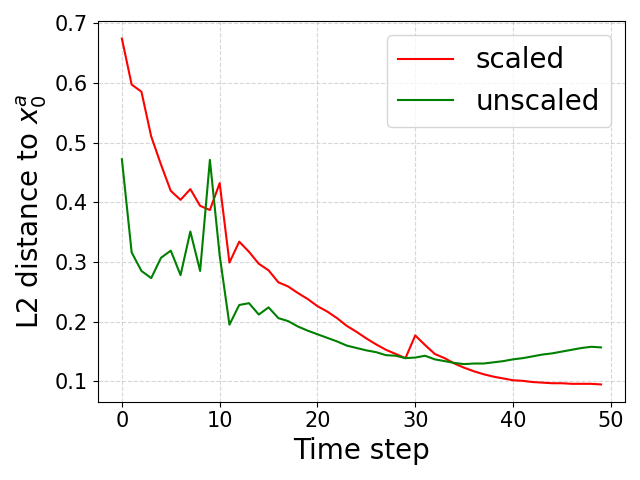} 
\caption{\textbf{Numerical comparison of the predicted $\mathbf{x}_0$ in all time steps.}}
\label{supp_loss}
\end{figure}

We provide the full results of training the predicted $\mathbf{x}_0$ in Fig.~\ref{supp_predx}. The right diffusion process without the time-aware scaling will first approach the ideal output and then deviate from the ideal output. The corresponding $L_2$ distance between the predicted $\mathbf{x}_0$ at every time step is shown in Fig.~\ref{supp_loss}. Empirically, we conclude that this is due to equally penalizing the distance to ideal output at every time step and the distance in early steps is very large. Without scaling, the large penalty on the early steps may influence the fitting on the later steps. After scaling, the model cares more about the final output so that the final output is close to the ideal output. Based on our experiment results, such a pattern is a common phenomenon during the training process.

%% file: secs/supp_quantitative.tex
\clearpage
\section{Evaluation details of quantitative analysis and user study}
\label{impl}
\noindent\textbf{Implementation details.} Our model is based on the SD v1.5 which is also used in InsP2P to generate triplets of text and image pairs for retraining. We use the AdamW optimizer with the learning rate $\gamma = 0.001$ for all experiments. Similar to the textual inversion~\citep{gal2022image}, the text embedding $\mathbf{c}$ is initialized with a coarse descriptor of the after-image, and the number of corresponding tokens of $\mathbf{c}$ is selected based on this descriptor. We use the default CFG scale $w = 7.5$ in SD v1.5. We set the attention timestamp $\tau = 0.7$ and the weighted-scale $\beta(t) = e^{t-T}$ which exponentially increases with the time step $t$. For all experiments, we use one pair of visual prompts, adopt DDIM with $T = 50$ steps, and optimize the framework for $N=50$ epochs, which takes around 20 minutes in one Nvidia V100. We also noticed that for some visual prompts, fewer epochs can achieve the same results.

\noindent\textbf{Evaluation metrics.} We follow the evaluation metrics used in previous image editing methods~\citep{ju2023direct, nguyen2024visual}. We evaluate the ability to preserve the invariance of the original image, and the agreement of editing direction between the test and train image pair. For evaluation metrics, we evaluate the PSNR, LPIPS, SSIM, Visual-CLIP (V-CLIP), and Image-CLIP (I-CLIP). Besides, we further use DINOv2 to extract image features and calculate the same score. Concretely, the PSNR represents the similarity to the ground truth edited image. The SSIM indicates the structural similarity to the ground truth. The LPIPS represents the similarity in terms of the features extracted by the neural network. For editing quality, we calculate the Visual CLIP similarity~\citep{nguyen2024visual} which measures the cosine similarity between the before/after training and test pairs. The higher V-CLIP similarity indicates the editing direction is more similar to the transformation represented by the visual prompt. The I-CLIP measures the cosine similarity between the original and edited images in CLIP embedding space. The same procedure applies to the DINO score. The mathematical formulations are summarized as follows.
$$ 
CLIP_{<visual>} = cosine<\mathbf{x}^a_0 - \mathbf{x}^b_0, \mathbf{y}^a_0 - \mathbf{y}^b_0>
$$
$$ 
CLIP_{<image>} = cosine<\mathbf{y}^b_0, \mathbf{y}^a_0>
$$

\noindent\textbf{User study setting.} The user study evaluates the human preference for fidelity and editing analogy of edited images. We report the average scores of each part from 60 participants. Each participant was asked to answer 8 questions. For each question, the participant is asked to evaluate the editing results from two perspectives with a score from 5 (high) to 1 (low): 1) The editing analogy to the visual prompts. This indicates how closely the edited image follows the editing effect demonstrated by the visual prompt.  2) The fidelity to the original test image. This reflects if the edited image preserves the content that should not be manipulated. The results validate that our method ranks first for both fidelity and editing analogy to the visual prompts. A screenshot of the question is shown in Fig.~\ref{user}. In total, we collected 480 answers from our participants.

\NewTColorBox{Context_Box}{ s O{!htbp} }{%
  floatplacement={#2},
  IfBooleanTF={#1}{float*,width=\textwidth}{float},
  colframe=gray!50!black,colback=gray!10!white,title=Context,
  }

\NewTColorBox{SC_TIE_Box}{ s O{!htbp} }{%
  floatplacement={#2},
  IfBooleanTF={#1}{float*,width=\textwidth}{float},
  colframe=yellow!50!black,colback=yellow!10!white,title=Visual and Image VIE Rating Prompt Template 
  }

\NewTColorBox{SC_CIG_Box}{ s O{!htbp} }{%
  floatplacement={#2},
  IfBooleanTF={#1}{float*,width=\textwidth}{float},
  colframe=green!50!black,colback=green!10!white,title=Image VIE Rating Prompt Template (Editing Fidelity)
  }

\noindent\textbf{VIE score based on GPT-4o.} VIE~\cite{ku2023viescore} recently revealed that GPT4-o can evaluate AI-generated images, which is highly correlated with the human evaluation result. Thus, VIE can be used as a metric to evaluate the quality and effectiveness of AI-generated images. Based on the VIE, we adopt the GPT-4o to evaluate the same two properties as the CLIP and DINO, editing direction toward the visual prompt and fidelity toward the original image. Concretely, we create two metrics, Visual-VIE (V-VIE) and Image-VIE (I-VIE), to measure the similarity of editing directions and the similarity to the original image. We present the prompt templates here to ask the GPT-4o to rate two properties from 5 (high) to 1 (low). We first provide the context to the GPT-4o and then provide the rules to evaluate the editing results. In Fig.~\ref{gpt}, we demonstrate a case of evaluating the first editing effect, \textit{Turn the cat into a Shiba Inu}, in Fig.~\ref{gpt}. We see that GPT4-o can provide reasonable evaluation comments in terms of \textbf{editing analogy} and \textbf{editing fidelity}.

\noindent\textbf{Evaluation dataset.} The evaluation dataset is constructed in a similar way as in the VII and InstructPix2Pix~\citep{brooks2023instructpix2pix}. We use the two latest image editing and customization datasets, PIE~\citep{ju2023direct} and Dreambooth~\citep{ruiz2023dreambooth}, and P2P~\citep{hertz2022prompt} to construct the before-and-image pairs. PIE is a dataset for image editing, which consists of 700 images in real and artificial scenes featuring ten distinct editing types. We only select the \textbf{real images} for evaluation. Dreambooth is a dataset for image customization, which consists of 30 different objects. Each object is represented by a number of images which sum up to 3000 images in total. We manually construct and filter 500 before-and-after image pairs with high fidelity and clear differences as the evaluation dataset.

\begin{Context_Box}[htb]
You are a professional digital artist. You will have to evaluate the quality and effectiveness of the AI-edited images based on the given rules.
You will have to give your output in this way (Keep your reasoning concise and short.):\\
\{\\
"score" : [...],\\
"reasoning" : "..."\\
\}\\
\end{Context_Box}

\begin{SC_TIE_Box}[htb]
 
RULES: \\
First of all, an exemplar image pair is provided: the first being the original image and the second being an edited version of the first. The difference of the exemplar image pair defines the image editing effect. Then, an evaluation image pair is provided: the first being a new original image and the second being an AI-edited version of the first. 

The objective is to evaluate how successfully the second image is edited as the effect demonstrated by the exemplar image pair.

Note that sometimes the two images might look completely different due to the failure of the image edit.\\
On a scale of 0 to 5: \\
A score from 0 to 5 will be given based on the success of the editing.
(0 indicates that the scene in the edited image does not follow the editing effect defined by the exemplar image pair at all. 5 indicates that the scene in the edited image follows the editing effect perfectly.) \\
A second score from 0 to 5 will rate the degree of overediting in the second image.
(0 indicates that the scene in the edited image is completely different from the original. 5 indicates that the edited image can be recognized as a minimally edited yet effective version of the original.)\\
Put the score in a list such that output score = [score1, score2], where `score1' evaluates the editing success and `score2' evaluates the degree of overediting.\\
\end{SC_TIE_Box}

\begin{figure*}[htb] 
\centering 
\includegraphics[width=1\textwidth]{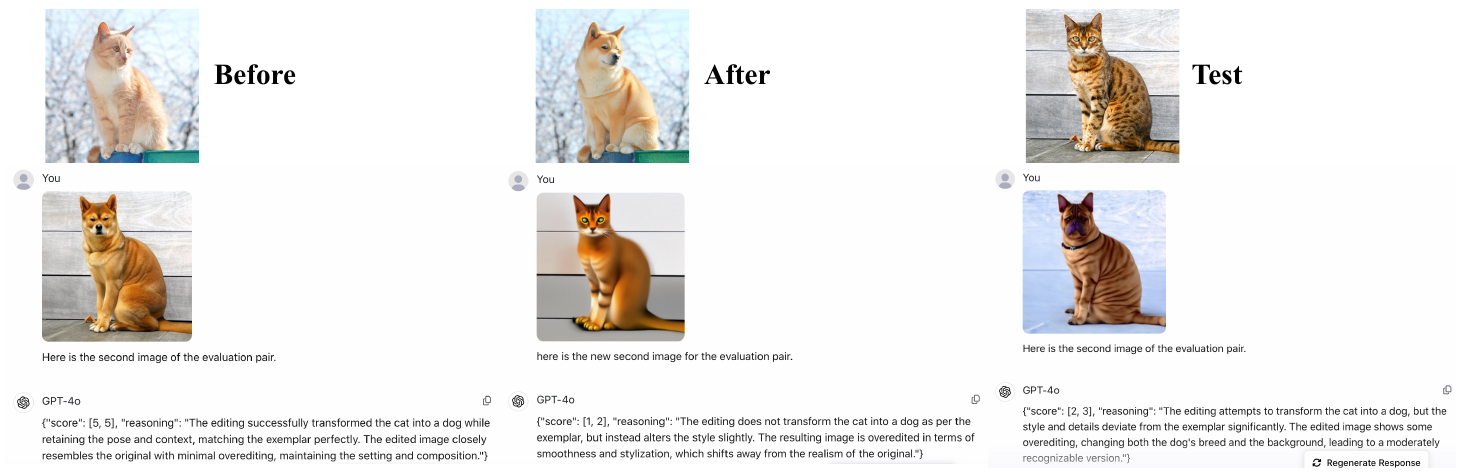} 
\caption{\textbf{Evaluation example of V-I-VIE scores from GPT4-o.} Following our prompt, we show GPT4-o can provide the reasonable and logical explanation for the evaluation score in terms of the editing analogy and fidelity.} 
\label{gpt}
\end{figure*}

\clearpage
\section{Additional qualitative editing results}
\label{add:qua}
We provide additional qualitative results to show that our method can edit different types of images which validates the generalization ability of our model. As shown in Fig.~\ref{supp_qua}, our model can learn different types of editing including style, texture, image tone, and changing objects. During the editing, our method can preserve the structure of previous methods and maintain high fidelity.

\noindent\textbf{Necessity of visual prompt.} In Fig.~\ref{supp_qua2}, we show the sub-styles of `painting' which cannot be straightly recognized and described by language alone. This demonstrates that visual prompts are necessary for image editing. Concretely, the ground truth editing prompt such as `neoclassic' and `psychedelic' cannot be easily identified. The comparison of the before-and-after image pair can convey accurate editing effects which are better than the single image. Besides, our method can also be used to edit objects such as changing cat to dog as shown in Fig.~\ref{supp_qua}.

\begin{figure*}[htb] 
\centering 
\includegraphics[width=1\textwidth]{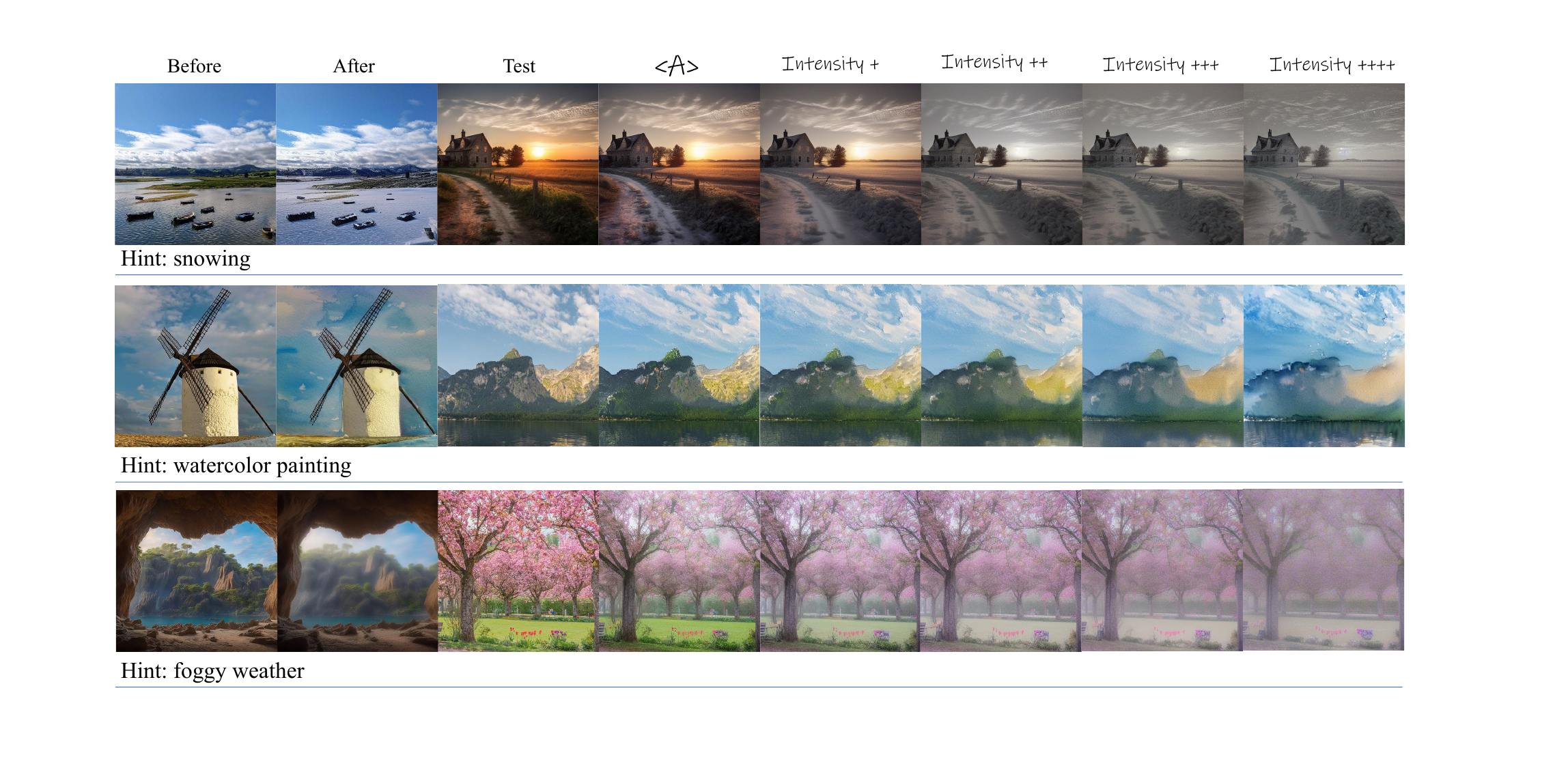} 
\caption{\textbf{Intensity control of editing effects from the visual prompt.} Our method can control the semantic intensity of the editing effect while preserving the structure information.} 
\label{inten}
\end{figure*}

\begin{figure*}[htb] 
\centering 
\includegraphics[width=1\textwidth]{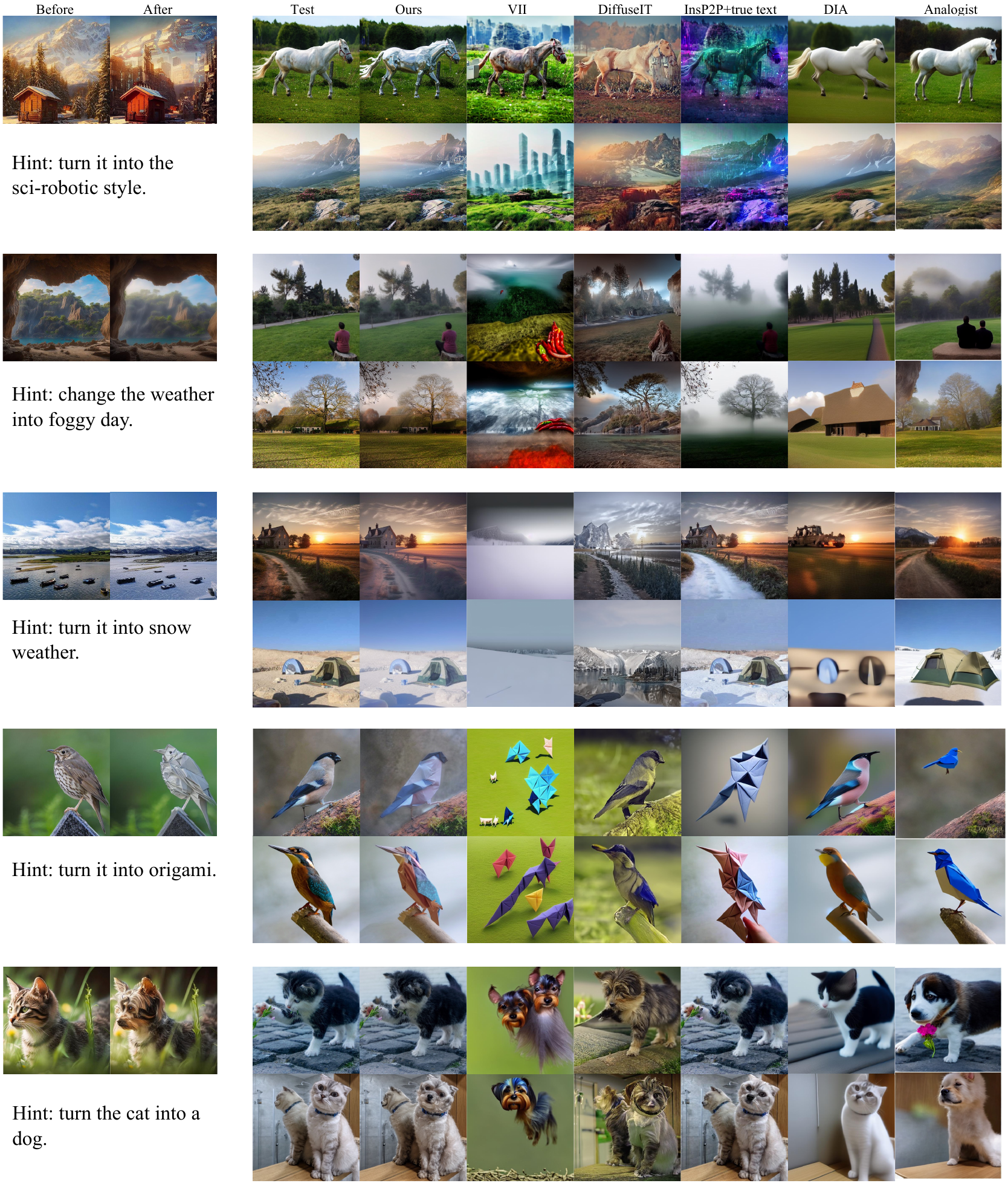} 
 \caption{\textbf{Additional qualitative results on real images.} We show extra editing results from different visual prompts. Our method can be better generalized to edit images from different scenes and domains.} 
\label{supp_qua}
\end{figure*}

\begin{figure*}[htb] 
\centering 
\includegraphics[width=1\textwidth]{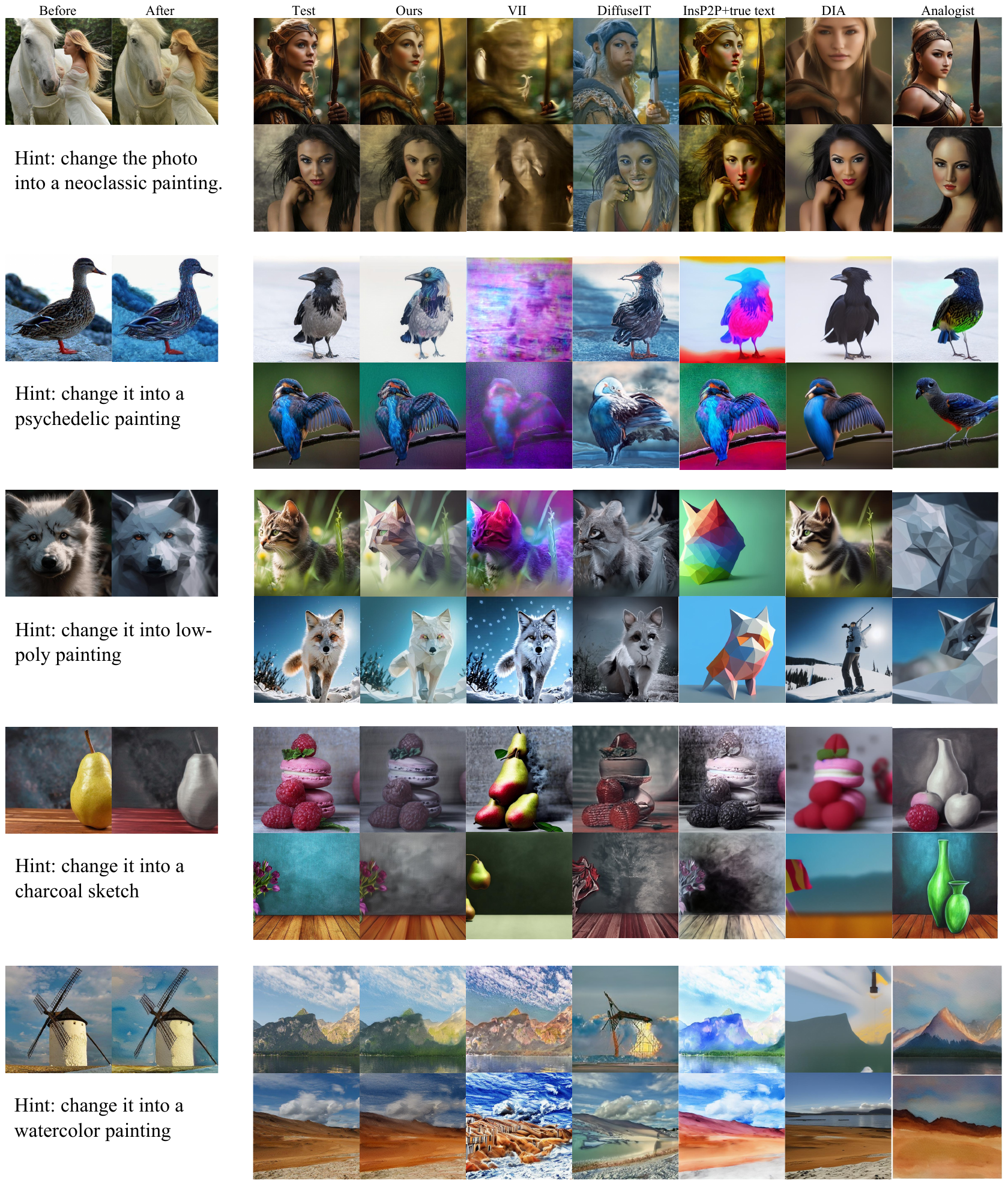} 
\caption{\textbf{Various sub-styles for `change it into a painting' on real images.} We show our method can learn various painting sub-styles, which are not intuitively recognized and described in language, from the visual prompt. This necessitates the visual prompt for image editing and validates our method's editing accuracy and generation ability. \textit{Please zoom in for details.}} 
\label{supp_qua2}
\end{figure*}

\noindent\textbf{Intensity control.} We also show the intensity control of the learned editing effects in Fig.~\ref{inten}. Except for the image tone change in Fig. 1, by changing the weight applied to the cross attention corresponding to the learned text embedding, our method can control the semantical intensity, such as snowing, foggy, and watercolor. Under different degrees of intensity, the structure and semantics of the original image can be preserved.

\begin{figure*}[htb] 
\centering 
\includegraphics[width=0.95\textwidth]{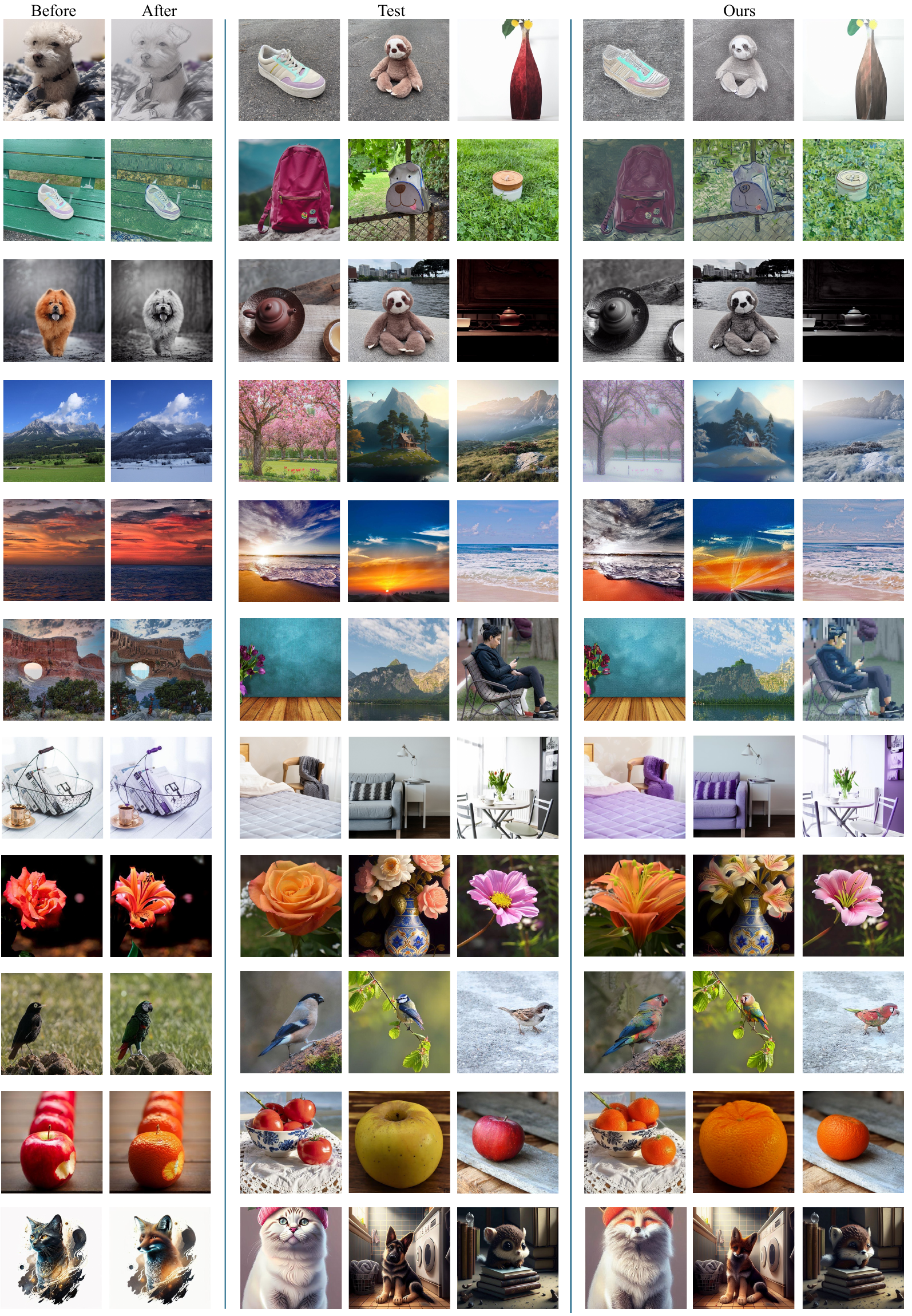} 
\caption{\textbf{Additional qualitative results of our method.}} 
\label{gpt}
\end{figure*}

\clearpage
\section{Study of failure case in limitation}
\begin{figure*}[htb] 
\centering 
\includegraphics[width=0.95\textwidth]{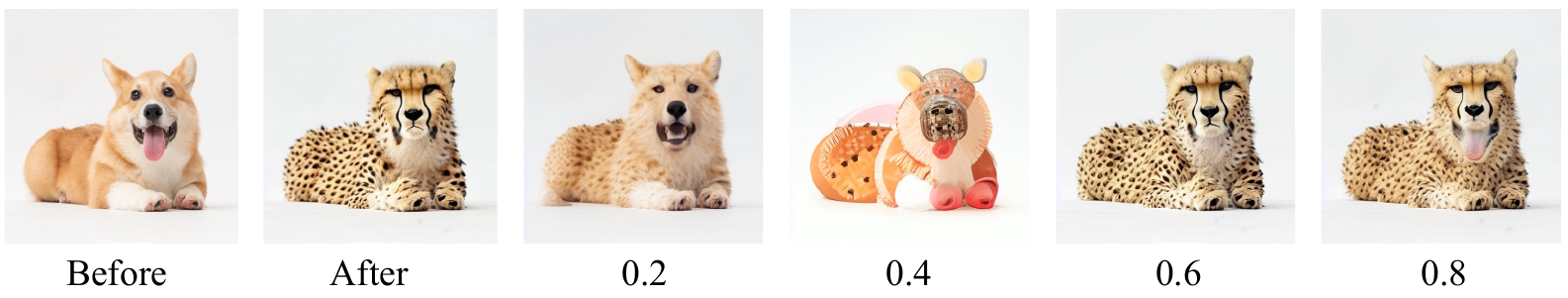}
\caption{\textbf{Training results with different attention injection timestep $\tau$.}}
\label{failure}
\end{figure*}

\begin{figure*}[htb] 
\centering 
\includegraphics[width=0.5\textwidth]{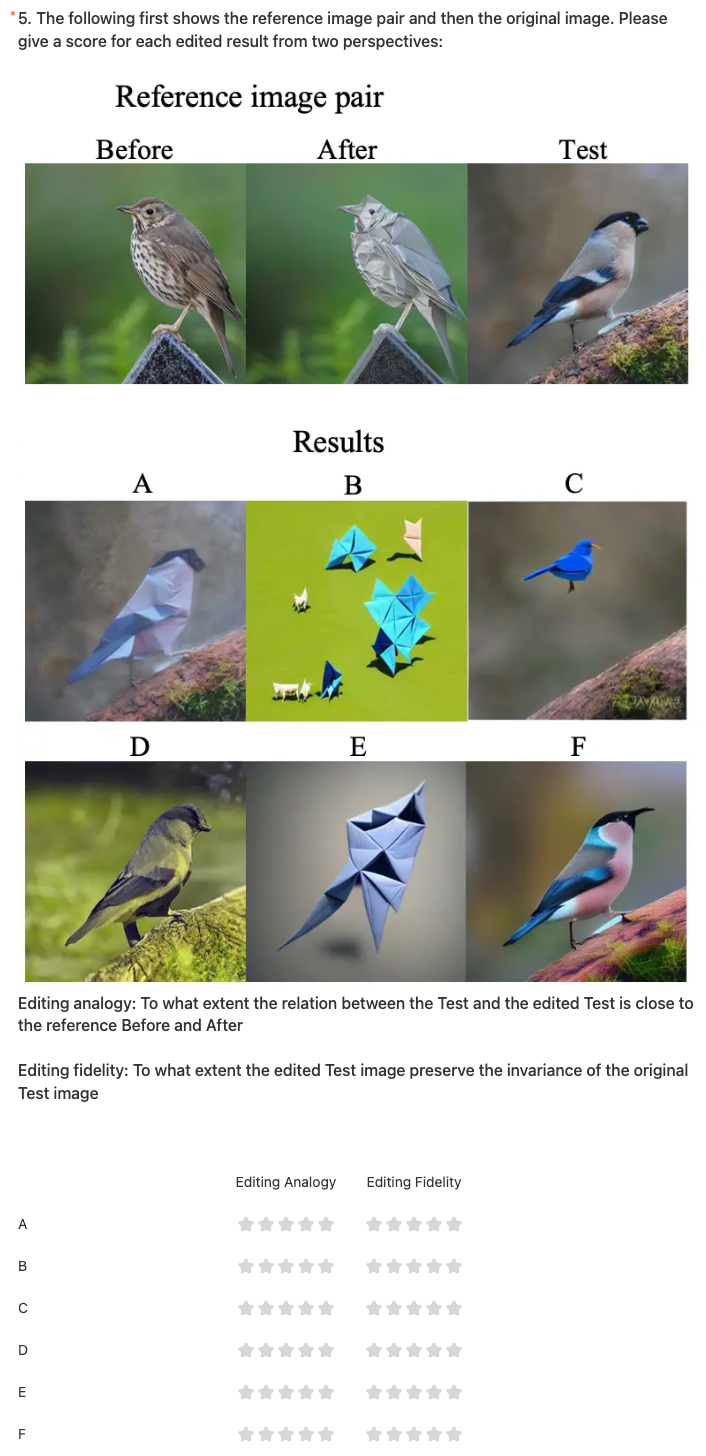} 
\caption{\textbf{Screenshot from the user study.} We evaluate the human preference of each result on editing fidelity based on the original image and the editing analogy regarding the visual prompt.}
\label{user}
\end{figure*}

Following the limitations addressed in the main paper, we present failure cases to analyze the influence of timestep $\tau$ on the editing effects. The timestep to inject the attention maps represents how much the source contents (the invariance) are inherited from the source image. Figure~\ref{failure} shows different $\tau$ can influence the training results. An appropriate $\tau$ can better minimize the training loss in Eq.13 and make training fit the target image. Otherwise, the training cannot perfectly fit the target image and will influence the editing results of new images.

\section{Quantitative ablation}
We further show the additional quantitative ablation to validate the effectiveness of the proposed modules. In Fig. 5, we already show that differential attention control is necessary to generalize to new images. Without it, the edited images are corrupted. This is consistent with the results in Table~\ref{qua_abla}. Besides, we also show that the time-aware scaling loss in Eq. 11 is effective in making the final output fit the target image.

\begin{table}[h]
        \centering
        \begin{tabular}{@{}l|c|c|c|c|c|c|c@{}}
        \toprule
            \textbf{Method} &\textbf{PSNR} &\textbf{SSIM} &\textbf{LPIPS} &\textbf{V-CLIP} &\textbf{I-CLIP} &\textbf{V-DINO} &\textbf{I-DINO}\\ \midrule
            w/o D-attn &3.21 &0.0601 &0.8123 &0.1136 &0.3213 &0.1027 &0.3663 \\ \midrule
            w/o scale &17.23 &0.5001 &0.4033 &0.1803 &0.7123 &0.1771 &0.7431\\ \midrule
            Ours &24.57 &0.8091 &0.1197 &0.2750 &0.8178 &0.3234 &0.9073 \\
            \toprule
        \end{tabular}
        \caption{\textbf{Quantitative ablation of differential attention and the time-aware scaling loss}. w/o D-attn: training without differential attention control. w/o scale: training without time-aware scaling loss.}
        \label{qua_abla}
\end{table}

